\documentclass[graybox]{svmult}

\usepackage{mathptmx}       
\usepackage{helvet}         
\usepackage{courier}        
\usepackage{type1cm}        
\usepackage{makeidx}         
\usepackage{graphicx}        
\usepackage{multicol}        
\usepackage[bottom]{footmisc}

\usepackage{amsmath}
\usepackage{cite}
\usepackage{url}
\usepackage{tikz}             
\usepackage{tikz-qtree}
\usetikzlibrary{matrix}
\usepackage[caption=false,font=footnotesize]{subfig}
\usepackage{float}
\usepackage{algorithm}
\usepackage{algpseudocode}
\usepackage{hyperref}
\usepackage{makecell}
\usepackage{graphicx}
\usepackage{adjustbox}

\usepackage{multirow}
\usepackage{algpseudocode}
\usepackage{colortbl} 
\usepackage{comment}

\makeindex             

\begin{document}

\title{Sharpness-Aware Minimization in Genetic Programming}
\author{Illya Bakurov†, Nathan Haut†, and Wolfgang Banzhaf}
\authorrunning{Illya Bakurov, Nathan Haut, and Wolfgang Banzhaf} 
\institute{Illya Bakurov \at Department of Computer Science and Engineering, Michigan State University, East Lansing, MI, USA \email{bakurov1@msu.edu}
\and Nathan Haut \at Department of Computational Mathematics, Science, and Engineering, Michigan State University, East Lansing, MI, USA \email{hautnath@msu.edu}
\and Wolfgang Banzhaf \at Department of Computer Science and Engineering, Michigan State University, East Lansing, MI, USA \email{banzhafw@msu.edu}
}

%
%
\maketitle


\abstract{Sharpness-Aware Minimization (SAM) was recently introduced as a regularization procedure for training deep neural networks. It simultaneously minimizes the fitness (or loss) function and the so-called fitness sharpness. The latter serves as a 
measure of the nonlinear behavior of a solution 
and does so by finding solutions that lie in neighborhoods having uniformly similar loss values across all fitness cases. 
In this contribution, we adapt SAM for tree Genetic Programming (TGP) by exploring the semantic neighborhoods of solutions using two simple approaches By capitalizing upon perturbing input and output of program trees, sharpness can be estimated and used as a second optimization criterion during the evolution. To better understand the impact of this variant of SAM on TGP, we collect numerous indicators of the evolutionary process, including generalization ability, complexity, diversity, and a recently proposed genotype-phenotype mapping to study the amount of redundancy in trees. The experimental results demonstrate that using any of the two proposed SAM adaptations in TGP allows (i) a significant reduction of tree sizes in the population and (ii) a decrease in redundancy of the trees. When assessed on real-world benchmarks, the generalization ability of the elite solutions does not deteriorate.}

\footnotetext[1]{† Authors contributed equally to this work.}


\section{Introduction}
\label{s1_intro}

The automatic discovery of mathematical expressions to describe phenomena captured in data is an extremely valuable tool for accelerating scientific discovery since the mathematical expressions can be used to make predictions about the systems that generated the data and the expressions can be directly studied to extract new insights into the system. There are many approaches for finding equations that fit data: linear regression, polynomial regression, SINDy~\cite{brunton2016discovering}, neural-symbolic regression~\cite{biggio2021neural}, symbolic regression~\cite{koza1994genetic}, etc. Genetic programming (GP) is a popular method for finding equations that fit data since it allows greater flexibility for the discovery of non-linear behaviors in data while also being effective in small data scenarios, unlike deep learning (DL) approaches which generally require large training data sets. This ability of GP to be effective in small data scenarios is likely in some part due to evolution's bias for simple solutions, and naturally simple solutions are less likely to overfit~\cite{Banzhaf2024}. Even so, in small data scenarios, the models are naturally underconstrained in the interstitial spaces between the training data points, which means that surprising and unexpected behavior can occur when interpolating. Ideally, we would want the models to be at least stable (smooth) when interpolating, otherwise trust in the models can be severely diminished. 

Some GP methods have been proposed to help lock down the behavior of models in these interstitial spaces to improve the robustness against overfitting in small data scenarios such as order of non-linearity~\cite{smits_nonlinearity_order}, model curvature~\cite{vanneschi2010measuring}, random sampling technique (RST)~\cite{random_sampling_technique}, RelaxGP~\cite{relaxGP}, and overfit repulsors~\cite{vanneschi2009using}. Order of non-linearity and model curvature are approaches that attempt to take properties of the model to predict if they are overfitting~\cite{smits_nonlinearity_order,vanneschi2010measuring}. Random sampling attempts to reduce the risk of overfitting by ensuring that no model sees the whole data set in a single generation~\cite{random_sampling_technique}. RelaxGP minimizes the risk of overfitting by assigning no additional fitness for getting closer than some threshold to the response data~\cite{relaxGP}. The use of overfit repulsors stores known overfit models and promotes the evolution of individuals that are different from the overfit models in semantic space~\cite{vanneschi2009using}. 

In this work, we propose using properties of the models to predict and discourage overfitting while also using synthetic data to catch models that are unstable in the interstitial spaces around the known training data. This approach is inspired by the Sharpness-Aware Minimization (SAM) technique recently developed in DL for neural networks~\cite{c21_sharpness_aware_dl_foret} and also applied in a different way in~\cite{zhang2024sharpnessaware} for feature construction. In~\cite{c21_sharpness_aware_dl_foret}, the authors demonstrate that SAM can promote generalization in neural networks. This is achieved by searching for neural network model parameters that lie in neighborhood of low sharpness. 

We introduce two new methods for selecting against sharpness in genetic programming: input-based SAM (SAM-In) and output-based SAM (SAM-Out). In this work, we aim to: 

\begin{itemize}
    \item Adapt sharpness-aware minimization from a deep learning regularization method to be compatible with genetic programming in an efficient way
    \item Reduce the risk of overfitting in GP by rewarding the smoothness of models rather than just the accuracy
    \item Improve the stability of models evolved using GP when interpolating by penalizing models that vary significantly in their response surfaces in neighborhoods in the fitness landscape. 
\end{itemize}
%
%
%
\section{Related Work}
\label{s2_related}
%
%
\subsection{Sharpness-Aware Minimization in Deep Learning}
\label{ss23_sam}
DL models commonly used in computer vision, are often comprised of hundreds of megabytes and millions of parameters, whereas large language models are made of billions of parameters and their size is measured in gigabytes. Training such complex systems is a big challenge for DL researchers and practitioners, and the high demand for computational resources and data availability are not the only constraints. These models are particularly prone to memorizing the training data rather than generalizing the learned patterns to unseen data (aka overfitting). Several techniques are commonly used to mitigate overfitting in DL and foster convergence: penalizing the fitness functions for disproportional high weights (such as L1 and L2 penalties), setting the weights connecting randomly selected neurons in some layer(s) to zero during a given training iteration, usage of varied data augmentation techniques to expose the model to more variations, normalization of batches of data that passes through the network, etc. 
Some DL researchers focused their attention on how a variety of neural architecture choices (such as depth, width, network architecture, optimizer selection, connectivity patterns, and batch size), affect the geometry of fitness landscapes and relate this to network generalization~\cite{c18_visualize_loss_nn_li}. H. Li et al. concluded that landscape geometry significantly affects the generalization ability of the system and that larger network depth, batch size and usage of sequential connections without shortcut connections produce sharper fitness landscapes and, consequently, negatively impact networks' generalization ability. 
Motivated by the connection between the fitness landscape and generalization, P. Foret et al. proposed a novel procedure that improves model generalization by simultaneously minimizing the loss value and sharpness of the loss landscape by promoting parameters that lie in the neighborhoods having uniformly low loss value~\cite{c21_sharpness_aware_dl_foret}.
Specifically, the authors add a regularization term to the loss function of the model that is a measure of how the training loss can be increased by \textit{moving} weights $W$ to a \textit{nearby} parameter value $W+\epsilon$, where $\epsilon$ represents the 
perturbation. The authors demonstrate that SAM improves model generalization ability across a range of widely studied computer vision tasks, and provides robustness to label noise on par with that provided by SOTA procedures that specifically target learning with noisy labels. This procedure was called Sharpness-Aware Minimization (SAM) and constitutes the main inspiration for the approach proposed in this manuscript.
%
%
\subsection{Semantic Awareness in Genetic Programming}
\label{ss21_gsgp}
Traditional crossover and mutation operators used in GP rely on 
structural (i.e., genotypic) transformations of parent individuals, without knowledge of their effects on the offspring's behavior. But a mere replacement of one node can lead to arbitrarily large changes in the behavior of a program (e.g., replacing $-$ with $*$ in $x_1+x2-x_3$). 

Since 2012 GP researchers have incorporated semantic awareness into GP. 
\footnote{The term {\it semantics} defines the vector of output values of a candidate solution (program), calculated on the training observations~\cite{c12_gsgp_moraglio}. Following this notion, a candidate solution in GP is a point in a multidimensional {\it semantic space}, where the dimensionality is equal to the number of observations in the training set.}
Specifically, Moraglio et al.~\cite{c12_gsgp_moraglio} proposed geometric semantic operators~(GSOs) that allow to transform genotypes of parents in such a way that the effect on the semantics of the offspring is known, unlike what happens with standard GP operators~\cite{b92_gp_koza}. 
A remarkable characteristic of GSOs is that the error surface on the training data for any SML problem is unimodal and therefore easy to search. This holds independently of the size and complexity of the dataset~\cite{b15_introduction_gsgp_vanneschi}.

In practical terms, GSOs allowed GP to achieve unprecedented levels of accuracy in numerous real-world applications, even when compared with other ML approaches~\cite{b14_gsgp_real_life_vanneschi,a21_stacking_gp_bakurov,c21_contemporary_sr_performance_lacava}.
Attracted by the characteristics of GSOs, many researchers focused on improving their effectiveness and efficiency. 
In the next paragraph, the reader can find some of the most prominent contributions that relate to one of the proposed SAM approaches in this manuscript.
Vanneschi et al.~\cite{b14_gsgp_real_life_vanneschi} demonstrated that using a sigmoid-bounding function in geometric semantic mutation (GSM) helps to stabilize the learning process and increase solutions' generalization. 
In this sense, the semantics of the offspring resulting from a GSM will surround the semantics of the parent within a user-controlled range $[-ms,\ ms]$, which can be interpreted as a step in any direction of the semantic space in a box of side $ms$.
Recently, Bakurov et al.~\cite{a24_norm_std_gsm_bakurov} proposed a normalization procedure for GSM that reduces the size of the programs and overcomes the saturation issues associated with the sigmoid function adopted in~\cite{b14_gsgp_real_life_vanneschi}, which resulted in more accurate and simple solutions. This was the GSM variant used in our study.
Gonçalves et al. introduced the concept of the semantic neighborhood - the set of neighbors reachable from a given solution when a GSM is applied to it - and proposed a stopping criteria based on neighborhood's properties~\cite{c17_semantic_neighborhood_goncalves}. Competitive generalization was achieved while using significantly fewer generations, and consequently in smaller solutions.
This suggests that information collected from the semantic neighborhood can bring notable value for the evolution.   
%
%
\subsection{Noisy Data and Fitness Functions}
\label{ss22_noisy}
Noisy fitness functions are utilized in GP to avoid the risk of overfitting. Rather than trying to find a "perfect" fit where no error is achieved, the goal becomes to find a "good-enough" fit such that the error is within some threshold. One such approach is RelaxGP~\cite{relaxGP}, where the authors change the fitness to instead consider an upper and lower bound for each response value rather than trying to perfectly fit each response value. 
%
%
\subsection{SAM in GP}
Very recently, sharpness was also incorporated in GP~\cite{zhang2024sharpnessaware}, in the scope of feature construction. In that work, sharpness was measured by introducing Gaussian noise to each node at all layers of the feature trees. The sharpness of an individual was then determined using the difference in loss after the noise was introduced. The results demonstrated that sharpness can indeed be used to reduce the size of GP trees and mitigate their overfitting tendency. One potential drawback of this approach is that it is computationally expensive since it computes sharpness using noise at each layer of the trees.
%
%
%
\section{Proposed Approach}
\label{s3_approach}

Here we propose two approaches for estimating the sharpness of a given tree in GP. 
Because of the fundamental differences 
between artificial neural networks (ANN) and TGP, 
a direct transfer of SAM from the former to the latter is impracticable. Typically, ANNs have a predefined (fixed) architecture (e.g., number of layers and nodes per layer) and the search for an optimal solution consists of finding an optimal set of weights (a collection of real values) that minimize the loss; therefore, for ANNs, the loss landscape is continuous. 
Although there were some successful attempts including learnable weights in TGP~\cite{c01_faster_gp_gradient_topchy_punch,c22_gsgp_hybrid_gradient_pietropolli,a23_gsgp_hybrid_gradient_pietropolli}, typically it is comprised exclusively of discrete structures (program elements), commonly divided into two groups: (i) terminals, which represent the input features of the problem and numeric constant values, and (ii) functions, which represent the operations to be performed on the terminals. The search for an optimal solution is conducted in a discrete space that is made of combinations of these structures in a tree data structure, that can grow or shrink dynamically in size. 
%
Given these differences, we use two approaches to adapt SAM for TGP. The first acts on the input of the tree by randomly perturbing both the constants of the tree and the input features of the model; in other words, it adds noise to the tree's terminal nodes. The second approach consists of random perturbation of the output of the tree. The subsections below explain each approach in detail.

\begin{figure}[h]
\centering
\caption{A sharp and smooth model is compared along with the values that were returned by our SAM-IN metric, showing that the sharper model is clearly identified by the metric.}
\label{fig:sharp_demo}
\includegraphics[width=10cm]{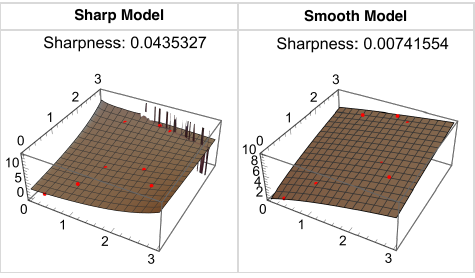}
\end{figure}

The goal of this approach is to identify and penalize models of the population that exhibit 
strongly nonlinear behavior in the fitness landscape (sharpness). 
An example of the behavioral difference of models is shown in Figure~\ref{fig:sharp_demo}, where we have two models with similar accuracy on the training data. However, one is not stable in some regions of the fitness landscape, while the other is very smooth and stable. The training points are shown overlayed on the model response surfaces and the sharpness, using our metric, is listed for each model in the labels above the plots. The values show that the sharpness metric would be able to identify this unstable model and remove it from the population. Unknowingly selecting and using the sharper model could be detrimental since it has very unstable behavior in certain regions of input space. 

\subsection{SAM on Input (SAM-IN)}
\label{ssec31_sam_in}
The first approach, called SAM-IN, quantifies the sensitivity of each tree in the population with respect to random perturbations in the terminals, both variables and constants. This sensitivity is calculated by measuring the absolute fitness differential after applying random noise of magnitude $\epsilon$ to a subset of input training data instances $n$ and all of the model constants. The noise introduced to the variables is a function of the standard deviation of the data in that variable. 
This quantity is then used as a second selection criterion during an evolutionary run. Selection is performed by using randomized double tournament selection, where the order of objectives is randomized for every selection. We expect that a large sensitivity towards input noise indicates that a given tree is likely overfit and, thus should have a lower selection preference when compared to a tree with a smaller sensitivity. Algorithm~\ref{alg_sam_in} contains an implementation of SAM-IN.

\begin{algorithm}
\caption{SAM on Input}\label{alg_sam_in}
\begin{algorithmic}[1]
\Require perturbations number $n > 0$, perturbation magnitude $\epsilon > 0$, training dataset $D$
\State $\overrightarrow{\sigma} \gets$ Compute the standard deviation of each feature in $D$   
\For{each generation in the evolutionary run}   
    \State $D_s \gets$ Select a random sample of size $n$ from $D$     
    \State $D_{s+\epsilon} \gets$ Make a copy of $D_s$
    \For{$i = 1$ to \Call{Length}{$D_{s+\epsilon}$}}
        \State $\epsilon_i \gets$ generate a random value in $\left [ -\epsilon\overrightarrow{\sigma},\  \epsilon\overrightarrow{\sigma} \right ]$
        \State $D_{s+\epsilon}[i] = D_{s+\epsilon}[i] + \epsilon_i$
    \EndFor
    \For{each tree $t$ in the population}
        \State $t_{\epsilon} \gets$ make a copy of $t$
        \For{each constant $c$ in $t_{\epsilon}$}
            \State $\epsilon_{c} \gets$ generate a random value in $\left [ -\epsilon,\  \epsilon \right ]$
            \State $c = c + \epsilon_{c}$
        \EndFor
        
        \State $f^{t} \gets$ compute the fitness of $t$ on $D_s$
        \State $f^{t+\epsilon} \gets$ compute the fitness of $t_{\epsilon}$ on $D_{s+\epsilon}$
        \State $f^{t}_{SAM-IN} \gets$ compute sharpness of $t$ as $|f_t - f_{t+\epsilon}|$
    \EndFor
\EndFor
\end{algorithmic}
\end{algorithm}

\begin{figure}
\centering
\caption{An example of a GP tree. The red arrows indicate the locations where noise is injected.}
\label{fig:tree_example}
\includegraphics[width=6cm]{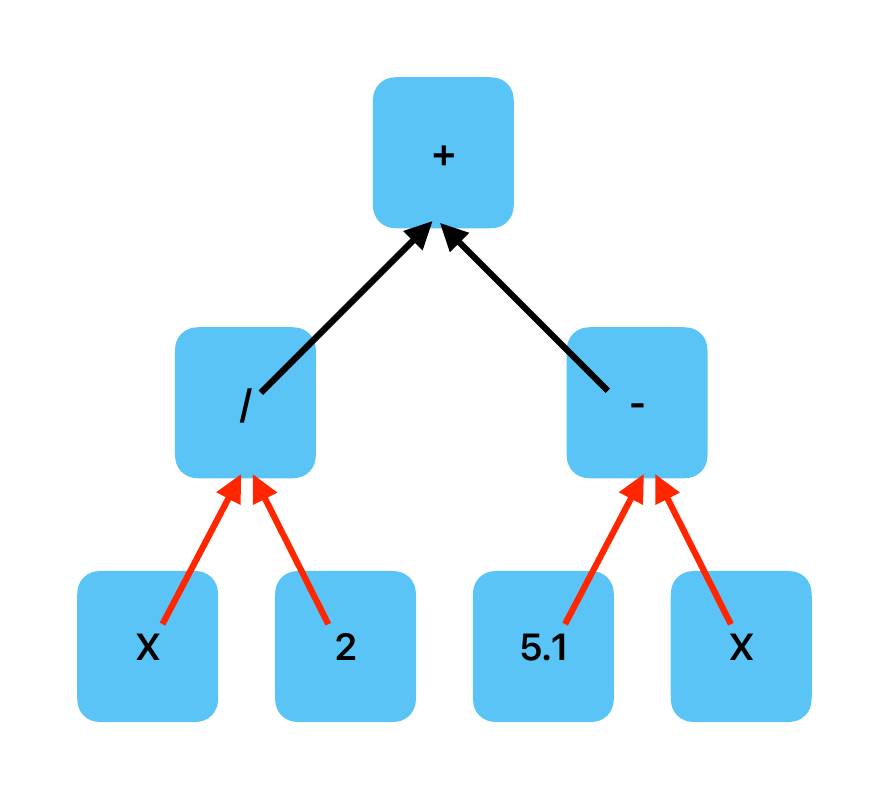}
\end{figure}

Unlike the original SAM in DL, or the SAM variant introduced in~\cite{zhang2024sharpnessaware}, we only introduce noise in the terminals rather than all of the connections in this method. This is demonstrated on a sample TGP in Figure~\ref{fig:tree_example}, which shows that noise is introduced only from the terminals as indicated by the red arrows. We did this under the assumption that the injected noise would propagate through all the layers of the tree and catch the sharp behavior. This is less computationally expensive than the original sharpness-aware approach, since we don't have to inject noise at each node of the tree. Further, since the goal of optimizing using sharpness as a metric is to produce more stable models, it is only necessary to detect sharpness with respect to noise in the terminals: Once a model is deployed, noise can only impact the terminals of the model.

\begin{figure}[h]
\centering
\caption{An example of using sharpness to generate a smoother approximation (left) of data compared to the original data generator (Ackley function - right). This would be useful in scenarios where the goal is to use the model to predict the location of a global optima.}
\label{fig:smoothness_demo}
\includegraphics[width=10cm]{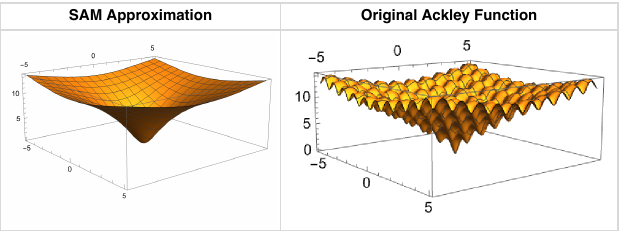}
\end{figure}

Another potential benefit of optimizing against sharpness is in scenarios where the goal is to use the model to predict some global optima, such as in active design of experiment. In those scenarios, developing a model with a smoother surface would make optimizing on the surface simpler and could possibly even lead to a convex surface which would make finding the global optima trivial. This is demonstrated in Figure \ref{fig:smoothness_demo}, where a sparse training data set was gathered from the 2D Ackley function and then GP using SAM-IN was used to generate a smooth model. The resulting model is very smooth, is convex, and has a global minima in the same location as the generating function. 

\subsection{SAM on Output (SAM-OUT)}
\label{ssec32_sam_out}
The second approach we examine, called SAM-OUT, builds upon the mechanics of the GSM operator and the notion of semantic neighborhood introduced by~\cite{c15_optimal_mutation_step_goncalves} and described in~\ref{ss21_gsgp}. Specifically, given a reference individual (tree), it generates $n$ semantic neighbors using GSM with $ms=\epsilon$, where $n$ and $\epsilon$ are hyperparameters of SAM-OUT. The fitness of each semantic neighbor is calculated and the sharpness of the reference individual is estimated as the variance in the semantic neighborhood. A large variance/instability in the semantic neighborhood indicates that a given reference individual is likely to overfit and, thus, should have a lower selection preference when compared to a tree with a smaller variance. Algorithm~\ref{alg_sam_out} contains an implementation of SAM-IN. Note that in this study, we use the normalized-GSM as proposed in~\cite{a24_norm_std_gsm_bakurov} given its notable stability and simplicity.

In practice, however, we do not need to construct and execute a random tree to sample from the semantic neighborhood of a reference individual, as the semantic neighbors are only used for the sharpness assessment. That is, the semantic neighbors will not be used anywhere else in the evolution. Thus, SAM-OUT, unlike SAM-IN, can fully operate on the semantic space, which makes it significantly more efficient. A direct implication of this is that one can generate a semantic neighbor by summing the semantics of the reference individual with a randomly generated vector of values centered at zero and bounded in $[-\epsilon,\ \epsilon]$ (which can be done in one line of code using NumPy or PyTorch libraries). 
It is important to note that there is no need for additional execution of the reference individual to extract its semantics, as it is already done when performing fitness evaluation.

\begin{algorithm}
\caption{SAM on Output}\label{alg_sam_out}
\begin{algorithmic}[1]
\Require perturbations number $n > 0$, perturbation magnitude $\epsilon > 0$, training dataset $D$
\For{each generation in the evolutionary run}
    \For{each tree $t$ in the population}
        \State $\widehat{y}_t \gets$ Compute the output of the tree $t$ on $D$
        \State $\sigma_{\widehat{y}_t} \gets$ Compute the standard deviation of $\widehat{y}_t$          
        \State $\overrightarrow{f^{t}_{n}} \gets$ empty vector for storing fitness of $n$ semantic neighbors of $t$        
        \For{each perturbation $i$ in $n$}
            \State $\widehat{y}^{i}_{t+\epsilon} \gets$ Make a copy of $\widehat{y}_t$ 
            \For{$j = 1$ to \Call{Length}{$\widehat{y}^{i}_{t+\epsilon}$}}
                \State $\epsilon^t_{(j, i)} \gets$ generate a random vector in $\left [ -\epsilon \sigma_{\widehat{y}_t},\  \epsilon \sigma_{\widehat{y}_t} \right ]$
                \State $\widehat{y}^{i}_{t+\epsilon}[j] = \widehat{y}^{i}_{t+\epsilon}[j] + \epsilon^t_{(j, i)}$
            \EndFor
            \State $\overrightarrow{f^{t}_{n}}[i] \gets$ compute the fitness of the $i^{th}$ semantic neighbor $\widehat{y}^{i}_{t+\epsilon}$    
        \EndFor
    \EndFor
    \State $f^{t}_{SAM-OUT} \gets$ compute sharpness of $t$ as $\sigma^2(\overrightarrow{f^{t}_{n}})$
\EndFor    
\end{algorithmic}
\end{algorithm}

As a result of using $R^2$ as opposed to RMSE as the fitness function, SAM-OUT is not functionally identical to introducing noise to the target vector, as would occur when using noisy fitness functions. This is shown in Table \ref{tab:sam_out} where the same error with RMSE is observed when adding the noisy vector, $\epsilon$, to the predicted target, $\hat{y}$, as when subtracting $\epsilon$ from the target $y$. This is, however, not the case when using $R^2$, indicating that adding noise to the output is different from adding noise to the target.

\begin{table}[h]
\caption{Comparison of SAM-OUT vs Noisy target for RMSE and $R^2$. $y$ represents the target variable, $\hat{y}$ represents the model predicted target, and $\epsilon$ represents a normally distributed noisy vector. $\epsilon$ was randomly initialized and then used in the same state for each evaluation in the table.}
\centering
\label{tab:sam_out}
\begin{tabular}{|l | c | c |}
 \hline
 \textbf{Noise Location} & \textbf{RMSE Value} & \textbf{$R^2$ Value} \\ [0.5ex] 
 \hline\hline
 $y+\epsilon$ & 2.931 & 0.431 \\
 \hline
 $y-\epsilon$ & \textbf{2.831} & 0.294 \\
 \hline
 $\hat{y}+\epsilon$ & \textbf{2.831}  & 0.817 \\
 \hline
 No Noise & 1.957 & 0.847 \\
 \hline  
\end{tabular}
\end{table}

\section{Experimental settings}
We assess our method on data from four real-world regression problems and four popular synthetic functions. The real-world problems are described in Table~\ref{table_data_real}. 
As for the latter, given a synthetic function $f(X)$, we randomly sample 100 data points in a 2-dimensional grid ($x$), under uniform distribution, where each dimension corresponds to an input feature ($x1$ and $x2$). The third dimension corresponds to the target of the prediction $y$ and is obtained as $f(x)=y$. The synthetic problems are described in Table~\ref{table_data_syn}. 
At each run, we use a different train-test partition of the data. For real-world problems, we randomly sample 70\% observations for training, while the remaining 30\% are held out for testing. For synthetically generated problems, we use 50\% for training and 50\% for testing.

\begin{table}[h]
\centering
\caption{Details about the four real-world datasets used.}
\label{table_data_real}
\begin{tabular}{|l | c | c | c |}
 \hline
 \textbf{Dataset} & \textbf{\#Instances} & \textbf{\#Features} \\ [0.5ex] 
 \hline\hline
 Boston~\cite{a78_housing_harrison} & 506 & 13 \\
 \hline
 Heating~\cite{a12_heating_load_tsanas} & 768 & 8 \\
 \hline
 Diabetes~\cite{a04_lar_diabetes_efron} & 442  & 10 \\
 \hline
 Concrete~\cite{misc_concrete_compressive_strength_165} & 1005 & 8 \\
 \hline  
\end{tabular}
\end{table}

\begin{table}[h]
\centering
\caption{Details about the four synthetic datasets used. In our experiments, $D=2$.}
\label{table_data_syn}
\resizebox{\columnwidth}{!}{  
\begin{tabular}{|l|l|c|}
\hline
\textbf{Name} & \textbf{Function}                                                                                                                                                    & \textbf{$x_i \in$}             \\ \hline \hline
Levy          &  \makecell{$f(\textbf{x}) =   \sin^2 (\pi w_1) + \sum_{i = 1}^{D - 1} (w_i - 1)^2 \left( 1 + 10 \sin^2 (\pi   w_i + 1) \right) + (w_d - 1)^2 (1 + \sin^2 (2 \pi w_d)) $, \\$w_i = 1 + \frac{x_i   - 1}{4}$} & {[}-10, 10{]}         \\ \hline
Ackley        &  \makecell{$f(\mathbf{x}) =   -a\;\exp\left(-b \sqrt{\frac{1}{D}\sum_{i=1}^D x_i^2}\right) -   exp\left(\frac{1}{D}\sum_{i=1}^D \cos(c\;x_i)\right) + a + \exp(1)$, \\$a=20$, $b=0.2$,   $c=2\pi$}      & {[}-32.768, 32.768{]}    \\ \hline
Rastrigin     & $f(\mathbf{x})   = 10D + \sum_{i=1}^D \left(x_i^2 -10cos(2pi x_i)\right)$                                                                                        & {[}-5.12, 5.12{]}     \\ \hline
Rosenbrock    & $f(\mathbf{x}) = \sum_{i=1}^{D-1} (100 (x_{i+1} - x_i^2)^2 + (x_i - 1)^2)$                                                                            & {[}-2.048, 2.048{]}   \\ \hline
\end{tabular}
}
\end{table}

Table~\ref{tab_hyperparameters_list} lists the hyper-parameters (HPs) used in this study, along with cross-validation settings. The HPs were selected following common practices found across the literature to avoid a computationally demanding tuning phase. 
$R^2$ was used as a fitness function~\cite{c03_gp_interval_scaling_keijzer,a13_plcc_livadotiotis,c23_gptp_r2_haut} as it was found to converge faster, generalize better, even when a few data points are available.
In this study, none of the operators is protected. Instead, programs that produced invalid values were automatically assigned a bad fitness value, making them unlikely to be selected. Although protecting operators will make sure that the function that the symbolic regression program will always return a numerical value on any input, leaving protected operators in would give the sharpness-aware approach an unfair advantage since sharpness would strongly bias against those models. Using unprotected operators where a model is given a fitness score of 1, the worst fitness, if a singularity is encountered moves the selection pressure against those solutions to the fitness metric rather than the sharpness metric, thus comparing sharpness-aware minimization and standard GP in a more fair setting.
No limit to tree depth was applied during evolution to allow bloat to reveal itself in all its \textit{splendor} and perform an unbiased assessment of how SAM can impact size growth and redundancy. 
To select individuals, two rounds of tournament selection are used (aka double tournament), with sizes 6 and 3 respectively. At each selection event, a criterion for selection is chosen at random (either training fitness or sharpness). The returned individual is the one that exhibits the highest $R^2_{train}$ and the smallest sharpness.
Recently, Banzhaf and Bakurov~\cite{arxiv24_gpm_banzhaf_bakurov} proposed an effective approach to map the genotype to the underlying behavioral determinants (aka phenotype) by removing semantically ineffective code from the former~\cite{arxiv24_gpm_banzhaf_bakurov}. Moreover, this genotype-phenotype mapping (GPM) is highly efficient as it is built upon the mechanics of the fitness evaluation (which inevitably needs to take place), avoiding, therefore, redundant function calls and calculations. 
This mapping shows that smaller phenotypes are hidden within larger genotypic trees, and can be easily extracted to facilitate interpretability. Also, they studied the population dynamics of both genotypes and phenotypes and concluded that the populations' behavior is normally based on a scarce number of unique and small phenotypes, which curiously happens even when evolution is more explorative. Additionally, the authors observed that the growth rate of phenotypes' size is notably smaller when compared to genotypes.
All of this suggests that using the GPM approach presented in~\cite{arxiv24_gpm_banzhaf_bakurov}, one can extract valuable insights about evolutionary dynamics. In this study, we use GPM to study whether sharpness-aware minimization influences the amount of unutilized code in evolved programs. 

To conduct our experiments we use the General Purpose Optimization Library~\cite{a21_gpol_bakurov}. GPOL is a flexible and efficient multi-purpose optimization library in Python that covers a wide range of stochastic iterative search algorithms, including GP. Its modular implementation allows for solving optimization problems, like the one in this study, and easily incorporates new methods. The library is open-source and can be found by following~\href{https://gitlab.com/ibakurov/general-purpose-optimization-library}{\underline{this link}}. The implementation of the proposed approach can be found there.

\begin{table}[t]
\centering
\caption{Summary of the hyper-parameters. Note that $P(C)$ and $P(M)$ indicate the crossover and the mutation probabilities, respectively.}
\begin{tabular}{|l|l|}
\hline
\textbf{Parameters}          & \textbf{Values}                                               \\ \hline\hline
\textnumero Train/test split & real-world problems: 70/30\%; synthetic problems: 50-50\%     \\ \hline
Cross-validation & Monte-Carlo (repeated random subsampling)                                 \\ \hline
\textnumero runs             & 60                                                            \\ \hline
\textnumero~generations      & 50                                                            \\ \hline
Population's size            & 100                                                           \\ \hline
Functions ($F$)              & \{+, -, x, /\, sin(x), cos(x), tanh(x), $x^2$, $x^{-1}$, $\sqrt{x}$, $e^x$, $log(x)$\}                                                \\ \hline
Initialization               & Ramped Half\&Half (RHH) with max depth of 5                   \\ \hline
Selection                    & double tournament with sizes 6 and 3, respectively            \\ \hline
Genetic operators            & \{swap crossover, subtree mutation\}                          \\ \hline
$P(C)$                       & 0.8                                                           \\ \hline
$P(M)$                       & 0.2                                                           \\ \hline
Maximum depth limit          & Not applied                                                   \\ \hline
Stopping criteria            & Maximum \textnumero~generations                               \\ \hline
SAM noise magnitude ($\epsilon$)  & \{0.1, 0.2, 0.5, 1.0\}                                   \\ \hline
SAM number of noisy perturbations ($n$) & \{10, 20, 50\}                                           \\ \hline
\end{tabular}
\label{tab_hyperparameters_list}
\end{table}
%
%
%
\section{Experimental results}
\label{sec5_results}
Table~\ref{tab_ranks} reports the ranks on the generalization ability obtained by different SAM variants and baseline TGP. To build the table, the $R^2$ fitness values on the test data observed in the last generation were taken and ranked in ascending order. Then, the ranks were averaged by algorithm type (standard GP and SAM variants). 
From the table, one can observe that several of the proposed SAM variants tend to rank the best more often than standard GP in terms of generalization ability. When comparing two distinct SAM approaches, SAM-IN ranks above its SAM-OUT counterpart. 
Additionally, from this table, we can extract the reference parameters for the two SAM variants. Both tend to produce better ranks using mild perturbations (0.1) and small neighborhood (10 and 20 for SAM-IN and SAM-OUT, respectively). From the table, these are $SAM-IN_{10}^{0.1}$ and $SAM-OUT_{20}^{0.1}$.
%
These SAM configurations were used in all figures of this section as the default SAM-IN and SAM-OUT, depicted as dashed dark blue and dark red lines. 

\begin{table}[]
\centering
\caption{Generalization ability ranks achieved by each version of SAM and standard GP. Large values correspond to high fitness ranks. The colors reflect the fitness: the green end of the green-yellow-red color scale highlights the highest (best) ranks observed at a experiment; the red end of the scale highlights the smallest (worse) ranks. The last row averages results of each column (SAM experiment).} 
\label{tab_ranks}
\resizebox{\columnwidth}{!}{  
\begin{tabular}{c||ccccccccccccc|}
\cline{2-14}
\multicolumn{1}{l|}{\textbf{}}     & \multicolumn{13}{c|}{\textbf{Algorithm}}                                                                                                                                                                                    \\ \hline
\multicolumn{1}{|c||}{\textbf{SAM}} & \multicolumn{1}{c|}{\textbf{GP}}                                    & \multicolumn{1}{c|}{\textbf{$SAM_{10}^{0.10}$}}   & \multicolumn{1}{c|}{\textbf{$SAM_{10}^{0.20}$}}   & \multicolumn{1}{c|}{\textbf{$SAM_{10}^{0.50}$}}   & \multicolumn{1}{c|}{\textbf{$SAM_{10}^{1.0}$}}    & \multicolumn{1}{c|}{\textbf{$SAM_{20}^{0.10}$}}   & \multicolumn{1}{c|}{\textbf{$SAM_{20}^{0.20}$}}   & \multicolumn{1}{c|}{\textbf{$SAM_{20}^{0.50}$}}   & \multicolumn{1}{c|}{\textbf{$SAM_{20}^{1.0}$}}    & \multicolumn{1}{c|}{\textbf{$SAM_{50}^{0.10}$}}   & \multicolumn{1}{c|}{\textbf{$SAM_{50}^{0.20}$}}   & \multicolumn{1}{c|}{\textbf{$SAM_{50}^{0.50}$}}   & \textbf{$SAM_{50}^{1.0}$}    \\ \hline\hline
\multicolumn{1}{|c||}{\textbf{IN}}  & \multicolumn{1}{c|}{\cellcolor[HTML]{E0E383}}                       & \multicolumn{1}{c|}{\cellcolor[HTML]{63BE7B}9.88} & \multicolumn{1}{c|}{\cellcolor[HTML]{A1D07F}8.63} & \multicolumn{1}{c|}{\cellcolor[HTML]{F0E784}7}    & \multicolumn{1}{c|}{\cellcolor[HTML]{FDD27F}6.13} & \multicolumn{1}{c|}{\cellcolor[HTML]{FEE282}6.5}  & \multicolumn{1}{c|}{\cellcolor[HTML]{F8696B}3.75} & \multicolumn{1}{c|}{\cellcolor[HTML]{FDD27F}6.13} & \multicolumn{1}{c|}{\cellcolor[HTML]{FBB078}5.38} & \multicolumn{1}{c|}{\cellcolor[HTML]{94CD7E}8.88} & \multicolumn{1}{c|}{\cellcolor[HTML]{BFD981}8}    & \multicolumn{1}{c|}{\cellcolor[HTML]{E4E483}7.25} & \cellcolor[HTML]{FDD27F}6.13 \\ \cline{1-1} \cline{3-14} 
\multicolumn{1}{|c||}{\textbf{OUT}} & \multicolumn{1}{c|}{\cellcolor[HTML]{E0E383}}                       & \multicolumn{1}{c|}{\cellcolor[HTML]{FA8F72}4.63} & \multicolumn{1}{c|}{\cellcolor[HTML]{B9D780}8.13} & \multicolumn{1}{c|}{\cellcolor[HTML]{FEE282}6.5}  & \multicolumn{1}{c|}{\cellcolor[HTML]{EAE583}7.13} & \multicolumn{1}{c|}{\cellcolor[HTML]{A7D27F}8.5}  & \multicolumn{1}{c|}{\cellcolor[HTML]{FA9573}4.75} & \multicolumn{1}{c|}{\cellcolor[HTML]{F0E784}7}    & \multicolumn{1}{c|}{\cellcolor[HTML]{F6E984}6.88} & \multicolumn{1}{c|}{\cellcolor[HTML]{FDD27F}6.13} & \multicolumn{1}{c|}{\cellcolor[HTML]{FEE282}6.5}  & \multicolumn{1}{c|}{\cellcolor[HTML]{FEE883}6.63} & \cellcolor[HTML]{FBAB77}5.25 \\ \cline{1-1} \cline{3-14} 
\multicolumn{1}{|c||}{\textbf{AVG}} & \multicolumn{1}{c|}{\multirow{-3}{*}{\cellcolor[HTML]{E0E383}7.38}} & \multicolumn{1}{c|}{\cellcolor[HTML]{E4E483}7.25} & \multicolumn{1}{c|}{\cellcolor[HTML]{ADD480}8.38} & \multicolumn{1}{c|}{\cellcolor[HTML]{FCEB84}6.75} & \multicolumn{1}{c|}{\cellcolor[HTML]{FEE883}6.63} & \multicolumn{1}{c|}{\cellcolor[HTML]{D8E082}7.50} & \multicolumn{1}{c|}{\cellcolor[HTML]{F97F6F}4.25} & \multicolumn{1}{c|}{\cellcolor[HTML]{FEE582}6.56} & \multicolumn{1}{c|}{\cellcolor[HTML]{FDD27F}6.13} & \multicolumn{1}{c|}{\cellcolor[HTML]{D8E082}7.50} & \multicolumn{1}{c|}{\cellcolor[HTML]{E4E483}7.25} & \multicolumn{1}{c|}{\cellcolor[HTML]{F3E884}6.94} & \cellcolor[HTML]{FCBE7B}5.69 \\ \hline
\end{tabular}
}
\end{table}

\begin{figure}[h]
\centering
\caption{Average population training (left) and test (right) fitness of SAM approaches vs. standard GP on the 4 real-world datasets.}
\label{fig_SAM_FitPop_lines_real}
\includegraphics[width=10cm]{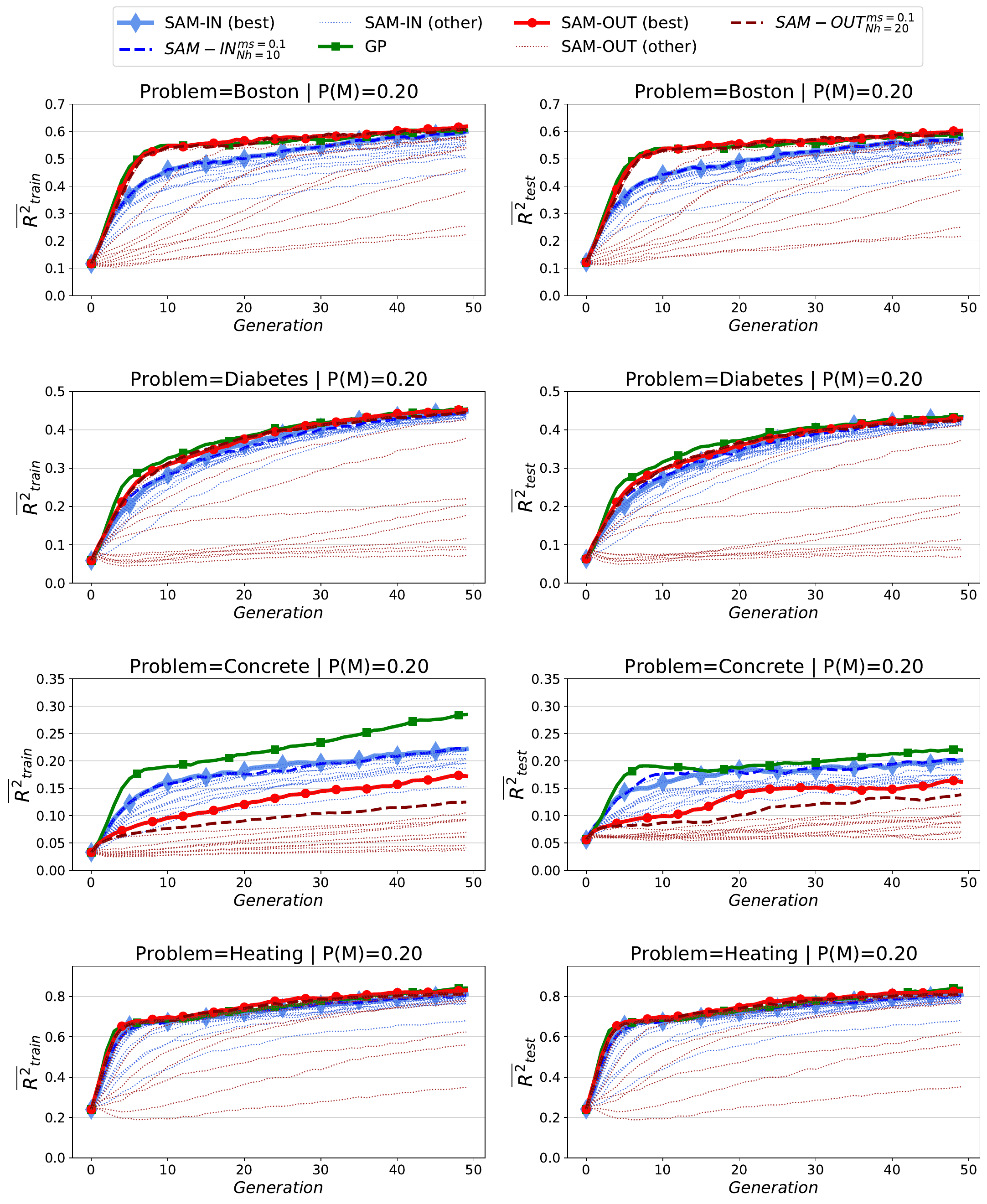}
\end{figure}

\begin{figure}[h]
\centering
\caption{Average population training (left) and test (right) fitness of SAM approaches vs. standard GP on the 4 synthetic datasets.}
\label{fig_SAM_FitPop_lines_syn}
\includegraphics[width=10cm]{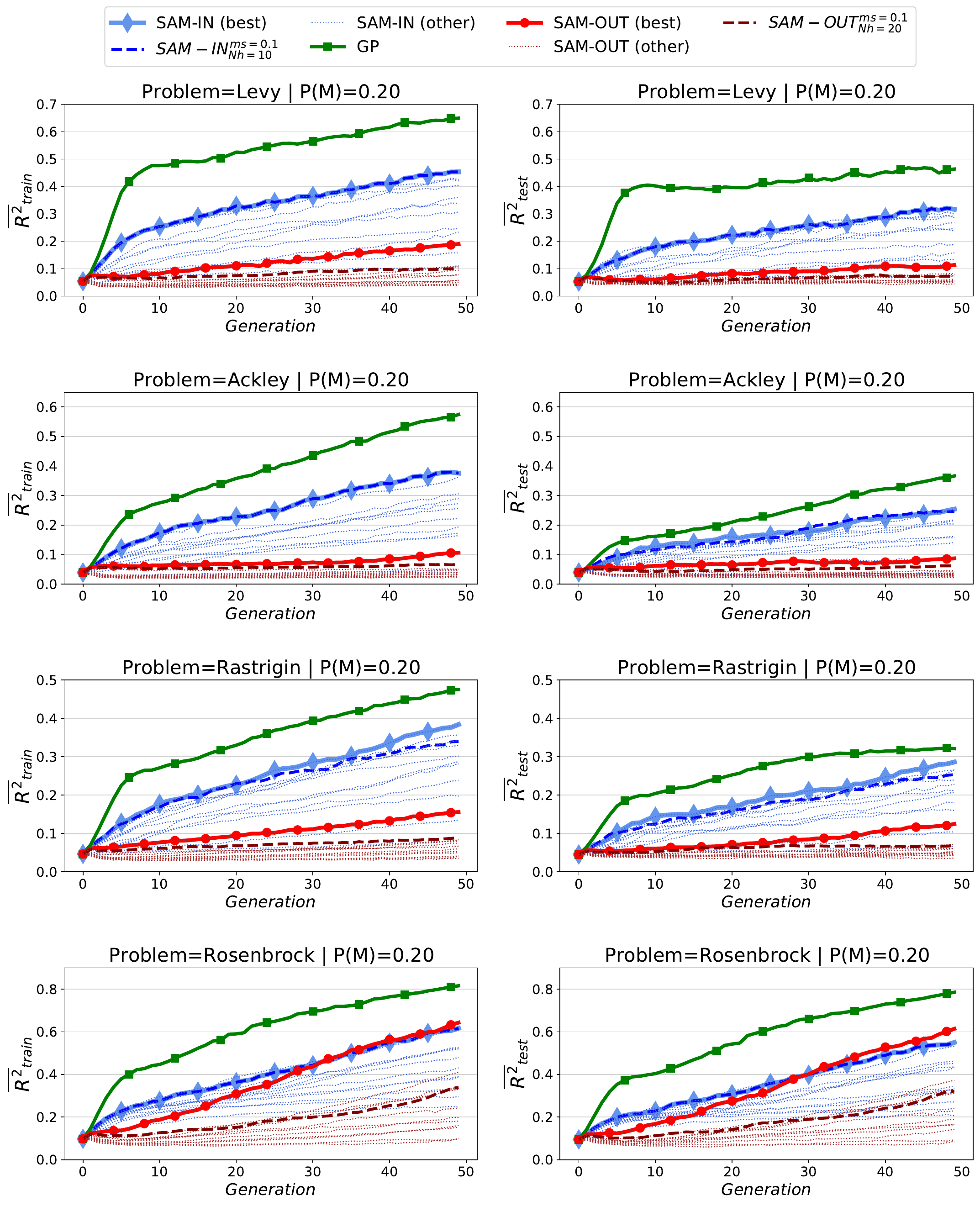}
\end{figure}

Figures~\ref{fig_SAM_FitPop_lines_real} and \ref{fig_SAM_FitPop_lines_syn} show the average fitness in the population for real-world and synthetic problems, respectively. Each problem is depicted in a given row, whereas each column represents a given partition: training on the left, and test on the right. 
The solid green lines represent standard GP (in green). The solid red and royal blue lines represent the SAM-IN and SAM-OUT methods that achieved the best generalization ability on a given problem. The SAM-IN and SAM-OUT configurations that were found to score the best more often across all the problems are depicted as dashed blue and dark red lines (these are the recommended SAM configurations and were obtained from Table~\ref{tab_ranks}). The remaining SAM-IN and SAM-OUT configurations are depicted as thin dotted lines, in blue and dark red, respectively. 
When looking at Figure~\ref{fig_SAM_FitPop_lines_real}, the average population fitness is roughly the same for standard GP and the SAM approaches in two problems: Diabetes Heating. On the Boston problem, SAM-IN shows a slower convergence, however, achieves similar fitness values at the end of 50 generations. On the Concrete problem, the highest fitness values are observed for standard GP, followed by SAM-IN and ultimately SAM-OUT. This particular problem seems more challenging to generalize on test data given the learning curves' instability. It is also relevant to point out that the recommended SAM configurations tend to overlap with the best SAM configuration for particular problems, highlighting their robustness across problems.
When looking at Figure~\ref{fig_SAM_FitPop_lines_syn}, we observe a higher average fitness in the population when using standard GP on all the problems, across both partitions. SAM-IN follows standard GP and SAM-OUT achieves the worst average fitness values on both partitions. This is not the case, however, on the Rosenbrock problem where the best SAM-IN and SAM-OUT achieve competitive performance. 

\begin{figure}[h]
\centering
\caption{Average best-performing model fitness of SAM approaches vs. standard GP on the 4 real-world datasets.}
\label{fig_SAM_FitElite_lines_real}
\includegraphics[width=10cm]{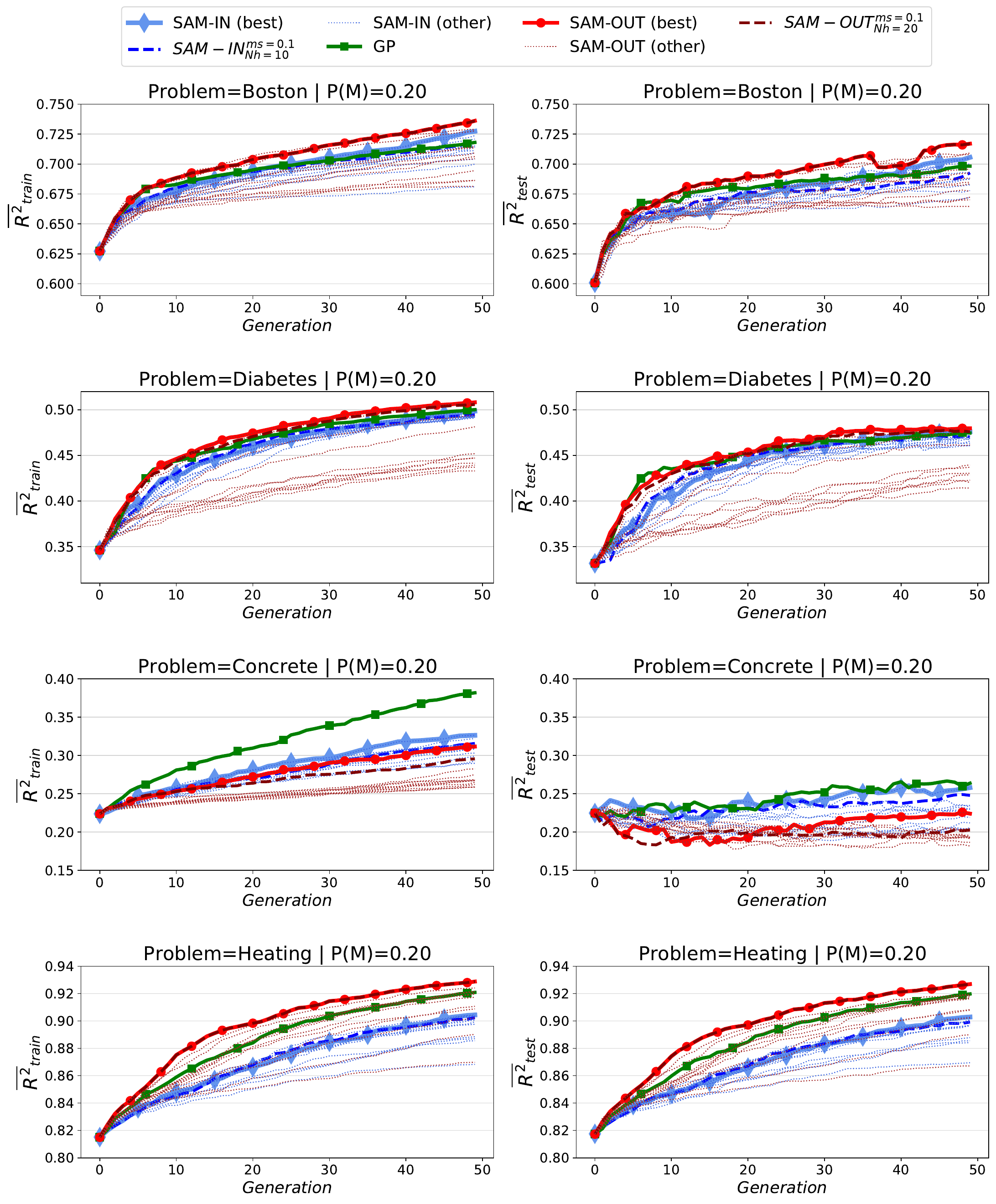}
\end{figure}

\begin{figure}[h]
\centering
\caption{Average best-performing model fitness of SAM approaches vs. standard GP on the 4 synthetic datasets.}
\label{fig_SAM_FitElite_lines_syn}
\includegraphics[width=10cm]{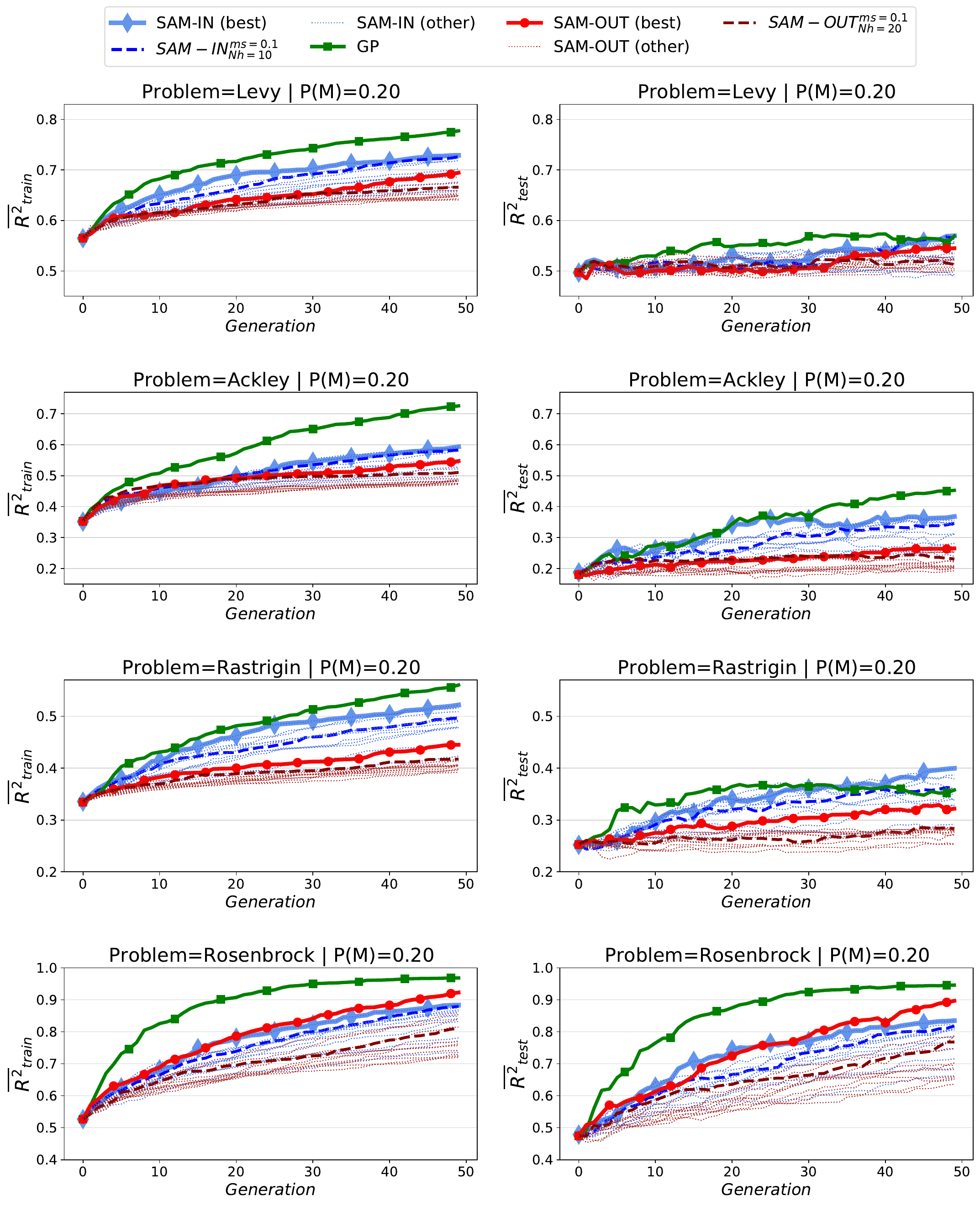}
\end{figure}

Figures~\ref{fig_SAM_FitElite_lines_real} and~\ref{fig_SAM_FitElite_lines_syn} show the training and test fitness of the elite individuals, averaged across the evolutionary runs. Figures' structure conforms to that of Figures~\ref{fig_SAM_FitPop_lines_real} and~\ref{fig_SAM_FitPop_lines_syn}.
When looking at Figure~\ref{fig_SAM_FitPop_lines_real}, the SAM-OUT converges faster and achieves better generalization when compared to both standard TGP and SAM-IN. Specifically, it achieves better test fitness than TGP on Boston and Heating problems, and similar fitness on Diabetes. Only on the Concrete problem, which seems to be challenging to generalize for all the algorithms,  SAM-OUT achieves worse generalization. 
SAM-IN produces similar results to TGP across three problems (Boston, Diabetes and Concrete), and scores the worst on Heating.
For the elite fitness on synthetic problems, depicted in Figure~\ref{fig_SAM_FitPop_lines_real}, standard TGP provided the best generalization, followed by SAM-IN which achieved the highest test scores on Rastrigin and similar scores on the Levy problem. SAM-OUT achieved the worst generalization ability, except on the Rosenbrock problem where it outperformed SAM-OUT but remained below TGP.

\begin{figure}[h]
\centering
\caption{Average length trees in the population (left) and the average amount of redundancy (right). The latter was calculated as the difference between the genotype and phenotype using the method proposed in~\cite{arxiv24_gpm_banzhaf_bakurov}. The figure regards 4 real-world datasets.}
\label{fig_SAM_PopLen_and_PopGPMDelta_lines_real}
\includegraphics[width=10cm]{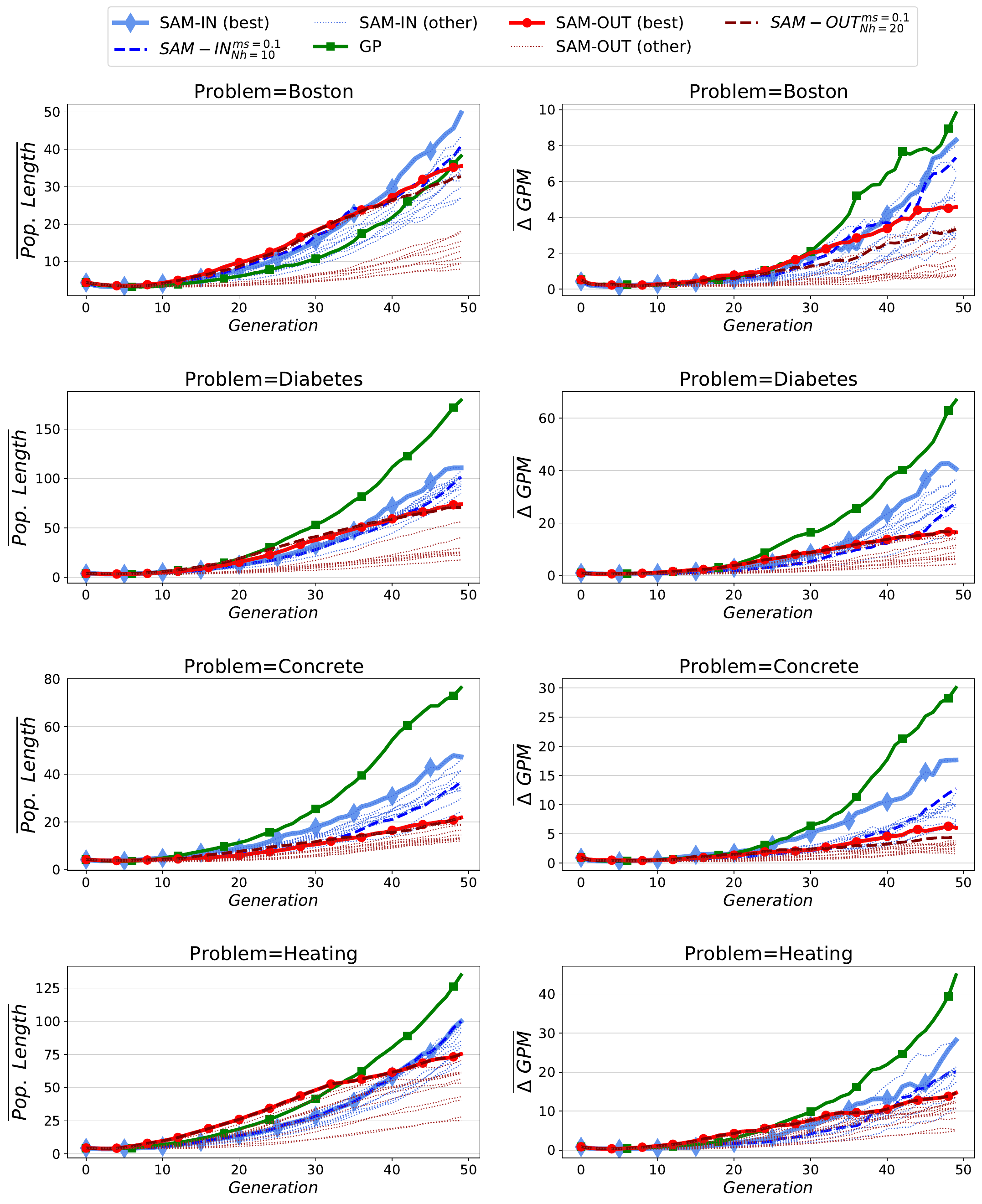}
\end{figure}

\begin{figure}[h]
\centering
\caption{Average length trees in the population (left) and the average amount of redundancy (right). The latter was calculated as the difference between the genotype and phenotype using the method proposed in~\cite{arxiv24_gpm_banzhaf_bakurov}. The figure regards 4 synthetic datasets.}
\label{fig_SAM_PopLen_and_PopGPMDelta_lines_syn}
\includegraphics[width=10cm]{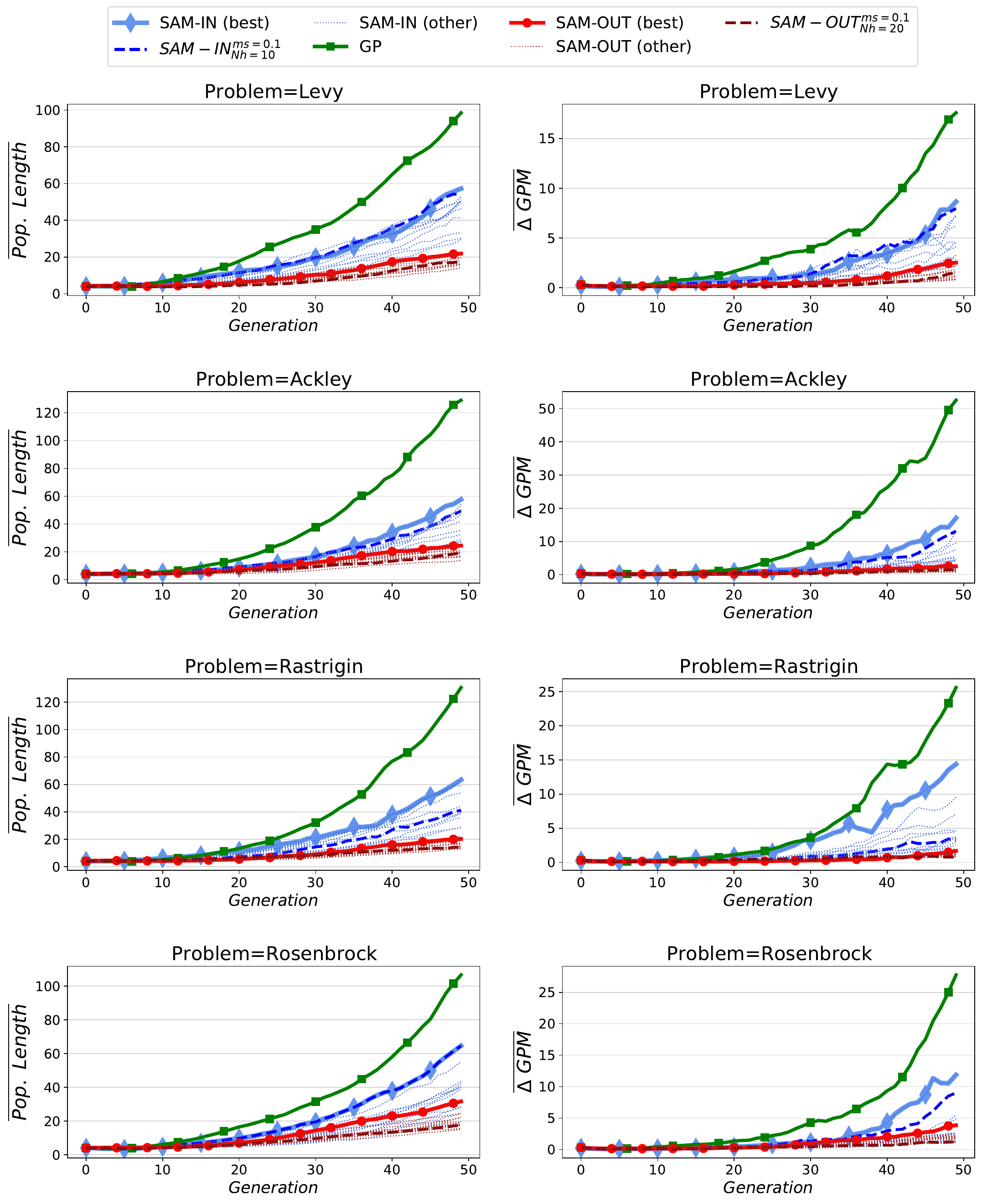}
\end{figure}

We also recorded the average length of trees (nodes' count) in the population across the evolutionary runs for all problems. Additionally, we measure the average difference between the whole genotype and the behavioral determinants (phenotype) of the individuals in the population using the GPM proposed in~\cite{arxiv24_gpm_banzhaf_bakurov}. The latter indicates the amount of redundant code (aka introns) in the trees. We display those results in Figures~\ref{fig_SAM_PopLen_and_PopGPMDelta_lines_real} and~\ref{fig_SAM_PopLen_and_PopGPMDelta_lines_syn} for the real-world and synthetic problems, respectively. The left column represents the average length, whereas the column on the right represents the average amount of redundancy. Each problem is depicted in a given row.
In a nutshell, the results show that both SAM methods tend to produce notably smaller trees and, within those trees, the amount of nodes that do not contribute to the behavior is significantly smaller (i.e., there is a higher utilization of the code). Usually, SAM-OUT produces smaller trees with less redundancy than SAM-OUT. Although this happens on both real and synthetic problems, it is particularly notable for the latter. 
For one real-world problem (Boston, Figure~\ref{fig_SAM_PopLen_and_PopGPMDelta_lines_real}), SAM-OUT reported larger average tree size. However, when looking at the amount of redundant code, it scores better than standard TGP.

\section{Conclusions}
In Machine Learning, the term \textit{learning} often refers to the task of inducing a general pattern from a provided set of examples, instead of \textit{memorizing} them. This ability is known as \textit{generalization} and represents one of the main battlegrounds in the field. 
In this study, we adapt the so-called Sharpness-Aware Minimization (SAM), previously introduced in Deep Learning (DL) to improve model generalization by simultaneously minimizing the loss and sharpness of the loss landscape by promoting parameters that lie in the neighborhoods having uniformly low loss values. 
To accommodate SAM in GP, we propose two methods. The first, called SAM-IN, measures sharpness by adding slight perturbations to the programs' terminals (i.e., both constants and input features), and penalizes for large fitness changes. The second, called SAM-OUT, measures sharpness by applying slight perturbations to the output of the programs (i.e., the semantics), and penalizes for large variance. In the latter case, sharpness is assessed without the need to execute the program, which makes it a computationally more efficient alternative. 

Comparing the two SAM approaches to standard GP, we observed that co-selecting individuals by fitness and SAM leads towards the evolution of smaller and less redundant models (i.e. models with a higher utilization of code) when compared to the standard fitness-only selection. For real-world problems, that are characterized by large dimensionality and noisy data, this was achieved without degradation of generalization ability. 

To minimize the computational overhead caused by SAM, we used a small number of perturbations (on both input and output). 
While this led to a faster evaluation of sharpness, it may have caused the SAM metric to be unstable compared to evaluating sharpness using the whole dataset since the metric may overlook localized sharpness in models if the subsample happens to select points only in smoother regions. Future work could explore the trade-off between subsample size and stability in the sharpness metric. 

Incorporating model sharpness as a metric to detect model overfitting and instability could be essential for improving the stability, generalization and simplicity of models developed by GP, leading to a better trustability. If a model becomes unstable when being applied to make predictions and returns values that are wildly unfeasible, the user would likely lose trust in the approach and seek out another method for developing models. Our method for detecting sharpness is also agnostic to the specific machine learning method being used since the perturbations only occur at the inputs and outputs of the models. This means that our approach is general and could easily be explored for any machine learning approach. We believe that the proposed sharpness metric should become part of the standard machine learning pipeline either during model training as a fitness metric or minimally as a post-processing selection criteria to ensure the models being deployed are stable.

\begin{acknowledgement}
WB acknowledges support from the Koza Endowment fund administered by Michigan State University. Computer support by MSU's iCER high-performance computing center is gratefully acknowledged.
\end{acknowledgement}


\bibliographystyle{spmpsci}
\bibliography{bibliography}

\end{document}